\documentclass[manuscript,screen, nonacm]{acmart} 

\AtBeginDocument{%
  }

\setcopyright{none} 
\newcommand{\xhat}{\hat{x}}
\newcommand{\dhat}{\hat{d}}
\newcommand{\yhat}{\hat{y}}

\newcommand{\shat}{\hat{s}}
\newcommand{\ahat}{\hat{a}}
\newcommand{\Pcal}{{\mathcal{P}}}
\newcommand{\sbar}{\bar{s}}
\newcommand{\R}[1]{\mathbb{R}^{#1}}
\newcommand{\E}[1]{\mathbb{E}[{#1}]}
\DeclareMathOperator{\subjectto}{subject~to}
\newcommand{\SPD}{{\tt SPD}}
\newcommand{\EOD}{{\tt EOD}}

\newcommand{\algrule}[1][0.2pt]{\par\vskip.5\baselineskip\hrule height #1\par\vskip.5\baselineskip}

\usepackage[linesnumbered,ruled,vlined]{algorithm2e}

\begin{document}

\title[Using Synthetic Data to Mitigate Unfairness and Preserve Privacy in Collaborative Machine Learning]{Using Synthetic Data to Mitigate Unfairness and Preserve Privacy in Collaborative Machine Learning}

\author{Chia-Yuan Wu}
\email{chw222@lehigh.edu}
\affiliation{%
  \institution{Lehigh University}
  \city{Bethlehem}
  \state{PA}
  \country{USA}
}

\author{Frank E. Curtis}
\email{frank.e.curtis@lehigh.edu}
\affiliation{%
  \institution{Lehigh University}
  \city{Bethlehem}
  \state{PA}
  \country{USA}
}

\author{Daniel P. Robinson}
\email{daniel.p.robinson@lehigh.edu}
\affiliation{%
  \institution{Lehigh University}
  \city{Bethlehem}
  \state{PA}
  \country{USA}
}







\begin{abstract}
In distributed computing environments, collaborative machine learning enables multiple clients to train a global model collaboratively. To preserve privacy in such settings, a common technique is to utilize frequent updates and transmissions of model parameters. However, this results in high communication costs between the clients and the server. To tackle unfairness concerns in distributed environments, client-specific information (e.g., local dataset size or data-related fairness metrics) must be sent to the server to compute algorithmic quantities (e.g., aggregation weights), which leads to a potential leakage of client information. To address these challenges, we propose a two-stage strategy that promotes fair predictions, prevents client-data leakage, and reduces communication costs in certain scenarios without the need to pass information between clients and server iteratively. In the first stage, for each client, we use its local dataset to obtain a synthetic dataset by solving a bilevel optimization problem that aims to ensure that the ultimate global model yields fair predictions. In the second stage, we apply a method with differential privacy guarantees to the synthetic dataset from the first stage to obtain a second synthetic data. We then pass each client's second-stage synthetic dataset to the server, the collection of which is used to train the server model using conventional machine learning techniques (that no longer need to take fairness metrics or privacy into account). Thus, we eliminate the need to handle fairness-specific aggregation weights while preserving client privacy. Our approach requires only a single communication between the clients and the server (thus making it communication cost-effective), maintains data privacy, and promotes fairness. We present empirical evidence to demonstrate the advantages of our approach. The results show that our method uses synthetic data effectively as a means of reducing unfairness and preserving privacy.
\end{abstract}



\maketitle

\section{Introduction}

\begin{figure*}[t]
\centering
\includegraphics[width=1\textwidth]{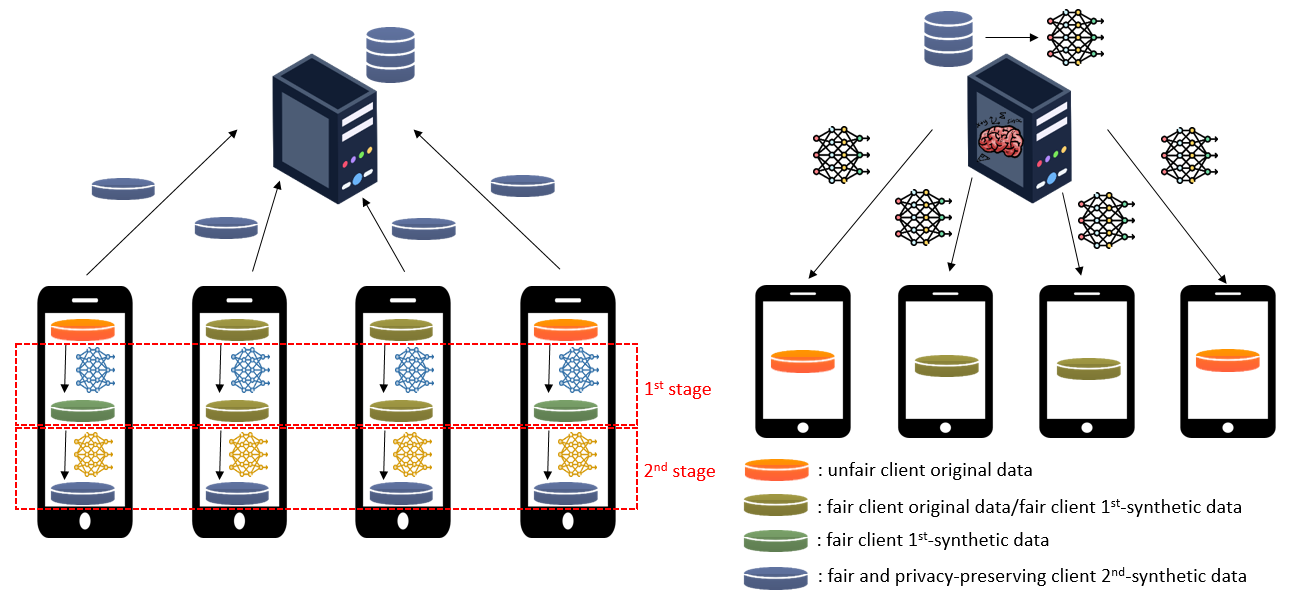} 
\caption{The above images illustrate our proposed approach for addressing fairness and privacy through collaborative machine learning. The images used in the illustration are from~\cite{flaticon}.}
\label{fig:idea_pic}
\Description{The figure includes two subfigures. The left side shows the actions of the clients. Each client learns its fair synthetic data from the client training data in the first stage, and then generates privacy-preserving synthetic data in the second stage using the first-stage synthetic data. The second-stage synthetic data is then passed to the server. The right side shows the server's action. The server trains a model on the combined client synthetic data, and then passes it back to all the clients.}
\end{figure*}

Fairness and privacy have become increasingly visible concerns as more machine learning (ML) tools are deployed in the real world. In the pursuit of highly accurate ML models, one often observes discriminatory outcomes when training is performed using unfair/biased data. Discrimination here refers to decisions made against certain groups determined by a sensitive feature such as race, age, or gender~\cite{mehrabi2021survey}. For example, racial bias has been identified in the commercial risk assessment tool COMPAS \cite{dressel2018accuracy} and in health systems \cite{obermeyer2019dissecting}, facial recognition technologies have shown accuracy disparities by gender and skin type \cite{buolamwini2018gender}, and age-related bias has been found in facial emotion detection systems \cite{kim2021age}. To mitigate such discriminatory practices, a collection of pre-processing, in-processing, and post-processing techniques have been studied in a data-centralized setting \cite{mehrabi2021survey, caton2024fairness, wang2021enhancing, zafar2017fairness, lohia2019bias, d2017conscientious}.  These techniques rely on access to the entire dataset, which in turn raises concerns about privacy leakage. Consequently, these approaches are not applicable to decentralized data environments such as in the context of Collaborative Machine Learning (CML) in the distributed setting.

CML is a special case of distributed learning that was originally introduced to distribute the computational workload across multiple machines to increase efficiency by using parallel computing resources. CML involves multiple learners working collaboratively to complete a task, e.g., train a server model. More recently, such learning has evolved to achieve additional objectives, such as learning directly on the datasets in multiple locations while keeping the training data private to preserve privacy \cite{laal2012collaborative, wang2022collaborative}. One of the most promising paradigms that prioritizes data privacy in CML is federated learning (FL) \cite{googleblog}. FL involves training client models using local data, sending the resulting client models to a server, and then learning a server model by aggregating the client models using an iterative process, thereby avoiding the need to have direct access to client data \cite{mcmahan2017communication}. Since the server model is indirectly derived from client data by using the client models as intermediaries, this can lead to good representation and generalization across all clients \cite{pan2023fedmdfg}. 

Separate from the typical distributed optimization setting, there are challenges in addressing fairness concerns in CML that include non-IID or unbalanced client datasets, massively distributed clients, and limited communication capacity between the clients and server \cite{mcmahan2017communication}. If some client datasets are inherently unfair, perhaps reflecting demographic or regional disparities, then this can lead to unfair server models after aggregation steps are performed. Ensuring fairness across multiple independent client models is difficult in this situation. Moreover, prior studies focusing on unfairness mitigation in CML have primarily relied on aggregation methods to obtain a fair server model. These approaches usually require frequent transmission of model-related information between clients and the server, which results in significant communication costs. Furthermore, additional client information, such as dataset size or unfairness measures, is needed to calculate the server model aggregation weights, which introduces potential risks of information leakage (i.e., a loss of privacy). Addressing these limitations is critical for advancing fairness in the CML context. 

To mitigate the aforementioned drawbacks, we propose a two-stage approach that only transfers \emph{synthetic} client datasets to the server instead of client data sets or client models.  The framework is illustrated in Figure~\ref{fig:idea_pic}. The left side shows the actions of the clients. Each client learns a synthetic dataset from its own training data with fairness considerations incorporated into the training process by solving a bilevel optimization problem in the first stage. The second stage involves using the first stage unfairness-mitigated synthetic data as input to a method that generates an additional data set with differential privacy guarantees. This step produces a second synthetic data set that ensures the privacy of the original data, which is then passed to the server. The right side of Figure~\ref{fig:idea_pic} shows the server's action. The server model is trained on a combined-from-all-clients (synthetic) dataset to minimize loss, just like training a typical data-centralized machine learning model without explicit fairness concerns. Since the client training data are preserved on the client devices, no additional information needs to be sent to the server for calculating model aggregation weights, and as the synthetic data sent to the server has differential privacy guarantees, this approach prevents any client information leakage. Moreover, our approach is designed as a one-time model transmission between the clients and the server. Once the server model is trained, the entire algorithm is completed when the server model is passed back to the clients, greatly reducing communication costs in certain scenarios.

We validate our approach empirically on datasets from the literature and summarize our contributions as follows:
\begin{itemize}
  \item To the best of our knowledge, this is the first CML approach that involves only a single transmission between the clients and server while also addressing unfairness issues. Thus, our approach avoids high communication costs while still producing a fair server model.
  \item A major advantage of our approach is that the server model can be trained without fairness concerns.  In a sense, the synthetic client data has already been generated to take fairness concerns into account, so all the server needs to do is aggregate the client data and train to maximize prediction accuracy.
  \item Our approach prevents potential client information leakage by using a single transmission of \emph{synthetic} data that has differential privacy guarantees from each client  to the server.
  \item Our proposed two-stage framework offers tremendous flexibility. First, one can directly control the amount of unfairness allowed through the choice of algorithm parameters (see $\epsilon^{SP}$ and $\epsilon^{EO}$ in~\eqref{prob:main.orig.client}).  Second, one can choose the sizes of the synthetic datasets transferred from the clients to the server (see $N_s^1$ and $N_s^2$ in Algorithm~\ref{alg:main}), thus explicitly controlling the communication cost.  Third, one can choose parameters for the second-stage so that a desired level of privacy is achieved (or skip the second-stage if one chooses to ignore privacy altogether).
\end{itemize}

\section{Background and Related Work}

In this section, we discuss background and related work on group fairness measures, fairness and privacy in distributed collaborative machine learning, bilevel optimization, synthetic data generation, and differential privacy.

\subsection{Group Fairness Measures}
Group fairness is a commonly used metric for determining whether a common group of people---as determined by a ``sensitive feature" (e.g.,  gender, race, or age)---is treated fairly. One of the prominent group fairness measures is \textit{statistical parity} (SP), also known in this context as \textit{demographic parity}. SP defines fairness as the equal probability of predicting positive labels for different sensitive groups, i.e., for groups characterized by the value of a sensitive feature. Another measure is \textit{equal opportunity} (EO), which requires the equal probability of correctly predicting positive outcomes conditioned on the true label being positive across sensitive groups \cite{pessach2022review}.

To define SP and EO concretely, let $X$ (resp., $S$) be a random vector representing nonsensitive (resp., sensitive) features and let $Y$ be a binary-valued random variable representing a label, where $Y \in \{-1, 1\}$, all defined over a combined probability space with probability measure $\mathbb{P}$. For convenience, let $A = (X,S)$. Let $y_\theta(\cdot)$ denote a binary classification function defined with weights $\theta$.  Then, as used in~\citet{caton2024fairness}, the binary classifier $y_\theta(\cdot)$ is said to have SP if and only if
\begin{equation} \label{equ:disparate_impact}
  \mathbb{P}(y_\theta(A) = 1 | S = 0) = \mathbb{P}(y_\theta(A) = 1 | S = 1),  
\end{equation}
and EO if and only if
\begin{equation} \label{equ:EO_impact}
  \mathbb{P}(y_\theta(A) = 1 | S = 0, Y = 1) = \mathbb{P}(y_\theta(A) = 1 | S = 1, Y = 1).  
\end{equation}

From a computational perspective, enforcing either equation~(\ref{equ:disparate_impact}) or (\ref{equ:EO_impact}) is challenging because the probabilities lead to nonconvex functions of $\theta$. \citet{zafar2019fairness} used the decision boundary covariance as an approximation of the group conditional probabilities. In particular, the decision boundary covariance is defined as the covariance between the sensitive feature and the signed distance between the features and the decision boundary.  Thus, letting $d_{\theta}(\cdot)$ denote this distance function corresponding to $\theta$, the decision boundary covariance with respect to SP is
\begin{equation} \label{equ:DBC}
\begin{aligned}
\text{Cov}(S, d_{\theta}(A)) 
= \E{(S-\E{S}) d_{\theta}(A)} - \E{S-\E{S}}\E{d_\theta(A)} 
\approx \frac{1}{N} \sum_{i=1}^N (s_i - \sbar) d_{\theta}(a_i)
\end{aligned}
\end{equation}
where $\E{S-\E{S}} = 0$ \cite{zafar2017fairness}, 
$\{a_i\}_{i=1}^N = \{(x_i,s_i)\}_{i=1}^N$ are samples of $\prod_{i=1}^N A \equiv \prod_{i=1}^N (X,S)$, and  $\sbar = (1/N)\sum_{i=1}^N s_i$.

In this study, we use a linear predictor, i.e., $d_{\theta}(a) = a^T \theta$. Using this choice, and keeping in mind our goal to obtain fair predictive models in terms of SP (resp., EO), we will present an optimization framework that enforces the constraint 

\begin{equation} \label{equ:DBC_constraint}
\left| \frac{1}{N}\sum_{i=1}^{N}(s_i-\bar{s}) a_i^T\theta \right| \leq \epsilon
\qquad \left(\text{resp., }  \left| \frac{1}{N}\sum_{i=1}^{N}(s_i-\bar{s}) \left( \frac{1+y_i}{2} \right)a_i^T\theta \right| \leq \epsilon \right)
\end{equation}
where $\epsilon$ is a threshold that determines the level of unfairness that is to be tolerated. To evaluate the classifier $y_\theta$ in terms of SP (resp., EO), it is common to use the difference $P(y_\theta(A) = 1 | S = 1) - P(y_\theta(A) = 1 | S = 0)$ (resp.,  $P(y_\theta(A) = 1 | S = 1, Y = 1) - P(y_\theta(A) = 1 | S = 0, Y = 1)$), which motivates the following measures of statistical parity difference and equal opportunity difference, respectively, defined over $\{a_i\}_{i=1}^N$:
\begin{equation}\label{equ:fairness-measures}
\begin{aligned}
\SPD{}
&= \frac{|\{i: y_\theta(a_i) = 1 \cap s_i=1\}|}{|\{i:s_i=1\}|} - \frac{|\{i: y_\theta(a_i) = 1 \cap s_i = 0\}|}{|\{i:s_i=0\}|} \ \ \text{and} \\ 
\EOD{}
&= \frac{|\{i: y_\theta(a_i) = 1 \cap s_i=1 \cap y_i = 1\}|}{|\{i:s_i=1 \cap y_i = 1\}|} - \frac{|\{i: y_\theta(a_i) = 1 \cap s_i = 0 \cap y_i = 1\}|}{|\{i:s_i=0 \cap y_i = 1\}|}. 
\end{aligned}
\end{equation}
The closer \SPD{} (resp., \EOD{}) is to zero, the closer the predictor $y_\theta$ is to achieving perfect fairness for these measures.

\subsection{Fairness and Privacy in Distributed Collaborative Machine Learning}

CML is a generic distributed framework for model personalization that allows multiple clients to work together towards a common goal, such as training a global model. CML systems are designed to facilitate learning without the need for centralized data, which inherently helps preserve data privacy \cite{googleblog}. FL is one of the most prominent distributed optimization frameworks in CML (proposed by \citet{mcmahan2017communication}) that is designed to address challenges related to data privacy and large-scale datasets. Given $K$ client objective functions $\{f_c\}_{c=1}^K$, the standard FL objective function is
\begin{equation} \label{prob:FL}
\begin{aligned}
\min_{\theta \in \R{n}} \
 &f(\theta), \ \ \text{ where } \ \ f(\theta) = \frac{1}{K} \sum_{c=1}^K f_c(\theta).
\end{aligned}
\end{equation}
Formulation~\eqref{prob:FL} minimizes the average ($f$) of the local client objectives ($\{f_c\}$) over the model parameter vector ($\theta$)~\cite{pan2023fedmdfg}. It is typical in ML that the client's objective function has the form $f_c(\theta) = \ell(\{x_{i,c}, y_{i,c}\}; \theta)$ for some loss function $\ell$.

Many CML algorithms require clients to produce parameter vectors $\{\theta_c\}_{c=1}^K$ that are passed to the server to obtain a solution estimate of~\eqref{prob:FL}.  For example, the FL algorithm FedAvg~\cite{mcmahan2017communication} computes the server model parameter vector as
\begin{equation}
\begin{aligned}
\theta = \sum_{c=1}^K \frac{n_c}{n} \theta_c,
\end{aligned}
\end{equation}
where $n_c$ is the number of data points for client $c$ and $n = \sum_{c=1}^K n_c$ \cite{mcmahan2017communication}.

Some previous studies have focused on how to aggregate the model parameters in a manner that may be able to obtain a fair server model. For example, FairFed computes each client's aggregation weight as the difference in a fairness performance metric between the client model and the server model \cite{ezzeldin2023fairfed}; FAIR-FATE incorporates momentum updates in the server model to prioritize clients with higher client fairness performance than the server model \cite{salazar2023fair}; and FedGAN and Bias-Free FedGAN generate metadata from client generators on the server to ensure that a balanced training dataset is input to the server model \cite{rasouli2020fedgan, mugunthan2021bias}. However, these algorithms rely on frequent communication of the model parameter values between the clients and the server. If client devices are slow or offline, the communication costs can be expensive and pose a significant limitation. Furthermore, in addition to the clients' model parameters, extra client information such as dataset size and the fairness measures on local models must be passed to the server to compute the aggregation weights, which can lead to potential leakage of client statistics \cite{ezzeldin2023fairfed}.

\subsection{Bilevel Optimization}
Bilevel optimization is characterized by the nesting of one optimization problem within the other.  One formulation of a bilevel optimization problem can be written as follows:
\begin{equation}
\begin{aligned}
\min_{x \in \R{n},\ y \in \R{m}} \
 &F_u(x,y) \\
 \ \subjectto \ \
&x \in X,\ y \in \arg\min_{y \in Y(x)} F_{\ell}(x, y),
\end{aligned}
\end{equation}
where $F_u$ and $F_{\ell}$ are the upper-level (or outer-problem) and lower-level (or inner-problem) objective functions, respectively.  Note that allowable values for the upper-level problem variables is implicitly defined by the lower-level problem \cite{savard1994steepest, giovannelli2024bilevel}. Due to the hierarchical structure, decisions made by the upper-level problem affect the outcomes of the lower-level problem~\cite{sinha2017review}.

In this paper, we address fairness issues by formulating a bilevel optimization problem over clients. Our inner-problem objective is to train a model with minimum loss from a given synthetic dataset. Our outer-problem objective is to generate a synthetic dataset (one for each client) as the input to the inner problem. Moreover, when utilizing the model obtained from the inner problem and evaluating it on client training data, certain fairness constraints are enforced. The details of our approach are presented in Section~\ref{sec:approach}.

\subsection{Synthetic Data Generation}

The first stage of our approach is inspired by data distillation, which was first introduced in 2018 \cite{wang2018dataset}. \citet{wang2018dataset} presented the concept of distilling knowledge from a large training dataset into a small synthetic dataset. The idea is that the model trained on the synthetic dataset can perform approximately as well as one trained on the original large dataset. This method makes the training process more efficient and reduces the computational cost of model updates.

Synthetic data can also lead to benefits in the CML setting. In CML, the cost of communicating model parameter values between the server and clients is a major concern, especially when the model structure is complex and the number of model parameters is large. \citet{goetz2020federated} and \citet{hu2022fedsynth} proposed transferring synthetic datasets to the server instead of the client models, where the size of the synthetic datasets were chosen to reduce the communication cost. The use of synthetic datasets helps in recovering the model update on the server to improve the server model updates \cite{goetz2020federated} or in recovering the client model directly to make server model aggregation more efficient \cite{hu2022fedsynth}.

The main contribution of these previous works is to make the training process efficient. They do not address concerns about fairness. In contrast, we propose an extension of the previous works that improves fairness between clients in CML. While learning synthetic data for clients, we simultaneously distill knowledge of their training data into the synthetic data and mitigate unfairness.  Consequently, when the model is trained by the server using the synthetic data from the clients, certain fairness criteria can  be satisfied when the model is evaluated on the client's original data. In addition, we can fully control the synthetic dataset size and the composition of each sensitive group and label. Therefore, when the synthetic data is sent to the server, we can help prevent client information leakage.

\subsection{Differential Privacy}

Differential Privacy (DP) is a mathematical technique that helps protect individual privacy in data analysis by ensuring that the inclusion or exclusion of a single data point has a negligible impact on a computational result (e.g., the outcome of a query function)~\cite{dwork2006differential, dwork2014algorithmic}. The widely used approach to implementing DP is to add calibrated random noise to the computation result. This injected noise makes the query outcome uncertain, and therefore makes it difficult to tell if an individual's data has been used (inherently preserving privacy).

To incorporate DP into ML, common approaches include introducing noise to gradients and clipping gradients when updating the model during the training process. ML models often learn the characteristics of the input training data, so these techniques prevent the model from memorizing specific individuals in the training data by limiting the impact of individual data on the model \cite{abadi2016deep}. A Generative Adversarial Network (GAN) \cite{goodfellow2020generative} is designed to generate synthetic data that resembles real training data, but it does not provide any privacy guarantee. To integrate DP into GAN, one can use the \textit{Post-Processing Theorem} \cite{dwork2014algorithmic} from DP.  In particular, if we use the differentially private discriminator (i.e., noise is added to the discriminator during training) to train the generator, then the generator becomes differentially private, and so does the synthetic data it generates \cite{jordon2018pate}.

In this paper, we propose a pipeline that includes the generation of differentially private synthetic data. We use an established differential privacy-guaranteed method (i.e., DP-CTGAN \cite{fang2022dp}) to accomplish this task. However, our framework is flexible enough to allow users to choose any synthetic data generation method they prefer, such as PATE-GAN \cite{jordon2018pate}, DPGAN \cite{ xie2018differentially}, or other suitable alternatives.  We stress that this aspect is only one part of our pipeline.

\section{Proposed Approach}\label{sec:approach}

In this section, we introduce our framework, which consists of two main parts: training synthetic datasets on the clients' side, and then training a global model on the server. For simplicity, we focus on the case with a single sensitive feature (taking values in $\{0,1\}$) and binary classification, although our approach could be generalized to other settings as well. 

\subsection{On the Client Side}\label{sec.client-side}

We address fairness concerns on the client side through the use of either SP or EO as our fairness measure and the covariance decision boundary as a relaxation. Each client learns an unfairness-mitigated and privacy-preserving synthetic dataset through a two-stage process. First, each client solves a bilevel optimization problem (see~\eqref{prob:main.orig.client} below) to obtain a fair synthetic dataset. Second, a differential privacy-based synthetic data generation method is applied using the first-stage synthetic data as input to create a privacy-preserving synthetic dataset. We now provide the details.


Consider an arbitrary client. Assuming the client has $N$ data points, for each $i \in \{1, ..., N\}$, we let $x_i \in \R{n-1}$ be the $i$th nonsensitive feature vector, $s_i \in \{0, 1\}$ be the $i$th  sensitive feature value, $y_i \in \{-1, 1\}$ be the label of the $i$th data point, $a_i = (x_i, s_i) \in \R{n}$ be the complete $i$th feature vector, and $d_i = (x_i, s_i, y_i)$ be the complete $i$th data point. The goal of each client is to learn a synthetic dataset of size $N_s^1$ with $N_s^1 \leq N$.  To denote the synthetic dataset, for each $i\in\{1,\dots,N_s^1\}$, we let $\dhat_i = (\xhat_i, \shat_i, \yhat_i)$ where $\xhat_i \in \R{n-1}$, $\shat_i \in \{0, 1\}$, $\yhat_i \in \{-1, 1\}$, and $\ahat_i = (\xhat_i, \shat_i) \in \R{n}$.

Multiple strategies can be used to define $\{(\shat_i, \yhat_i)\}$.  For example, if $N_s^1 = N$, then one can simply choose $\{(\shat_i, \yhat_i)\} =  \{(s_i, y_i)\}$.  On the other hand, if $N_s^1 < N$, then one can choose $\{(\shat_i, \yhat_i)\}$ so that the proportions of pairs (i.e., the pairs $(0,-1)$, $(0,1)$, $(1,-1)$, and $(1,1)$) are the same as those for $\{(s_i, y_i)\}$.  Details on how we choose these values for our numerical experiments are provided in Section~\ref{sec:experiments}.

Once $\{(\shat_i, \yhat_i)\}$ has been chosen, the ``ideal" problem solved by the client is defined, for user-defined tolerances $\{\epsilon^{SP},\epsilon^{EO} \}\subset [0,\infty)$ and loss function $\ell$, as follows:
\begin{equation}
\begin{aligned}\label{prob:main.orig.client}
\min_{\{\xhat_i\}} \ \
 &\ell\big(\{d_i\},\theta_{\text{ideal}}(\{\dhat_i\})\big) \\
 \ \subjectto \ 
 & \left|\frac{1}{N}\sum_{i=1}^{N}(s_i-\bar{s}) a_i^T\theta_{\text{ideal}}(\{\dhat_i\})\right| \leq \epsilon^{SP} \ \ \ \qquad\qquad \text{(SP constraint)} \\
 \text{or} \ 
 & \left|\frac{1}{N}\sum_{i=1}^{N}(s_i-\bar{s}) \left(\frac{1+y_i}{2} \right) a_i^T\theta_{\text{ideal}}(\{\dhat_i\})\right| \leq \epsilon^{EO} \ \ \ \text{(EO constraint)},
\end{aligned}
\end{equation}
where we define 
\begin{align*}
\theta_{\text{ideal}}(\{\dhat_i\})
&:= \arg\min_{\theta} \ \ell(\{\dhat_i\},\theta)
\ \ \text{and} \ \
\sbar := \frac{1}{N}\sum_{i=1}^{N}s_i.
\end{align*}
The loss function $\ell$ can be chosen based on the task performed. In this paper, since we focus on binary classification, we select the logistic loss function defined over data $\{d_i\} \equiv \{(x_i,s_i,y_i)\}$ and parameter vector $\theta\in\R{n}$ as
\begin{equation}\label{prob:logitloss}
\ell(\{d_i\},\theta)
:= \frac{1}{N} \sum_{i=1}^N \log(1+e^{-y_i a_i^T\theta}).
\end{equation}

The aim of problem~\eqref{prob:main.orig.client} is to learn synthetic nonsensitive features $\{\xhat_i\}$ in the synthetic dataset $\{\dhat_i\}$. The lower-level problem (or inner problem) is an unconstrained optimization problem that trains a model by minimizing the loss over the synthetic dataset. The upper-level problem (or outer problem) aims to minimize the loss of the model from the inner problem by evaluating it on the client's actual training dataset while satisfying fairness constraints.

Regarding the fairness constraints in~(\ref{prob:main.orig.client}), we want the covariance between the predictions and the sensitive features in the client training to fall within certain thresholds (i.e., $\epsilon^{SP}$ or $\epsilon^{EO}$). In this way, it keeps \SPD{} or \EOD{} close to zero to achieve good group fairness in terms of statistical parity or equal opportunity. 

In the second stage, the objective is to generate a second synthetic dataset of size $N_s^2$ that addresses privacy concerns (the first-stage synthetic dataset already addressed fairness). Any suitable privacy-preserving data generation method \cite{jordon2018pate, xie2018differentially, fang2022dp} can be applied here using the fair synthetic data from the first stage as input.  The resulting second synthetic dataset is both unfairness-mitigating and privacy-preserving, and can be passed to the server to train a global model.


\subsection{On the Server Side}

After each client learns its unfairness-mitigating and privacy-preserving synthetic dataset (the result of the second stage as described in the previous section), it is passed to the server. The server then trains a model on the combined synthetic dataset (i.e., the collection of synthetic datasets passed to it by the clients) defined using the same modeling formulation as in the client's inner optimization problem. Specifically, if we let $\dhat_{i,c}$ denote the $i$th data point for the $c$th client, then the server computes the model parameters for the global model by solving the problem
\begin{equation}
\begin{aligned}
\theta(\{\dhat_{i, c}\}) := \arg\min_{\theta} \ \ell(\{\dhat_{i, c}\},\theta).
\end{aligned}
\end{equation}
We stress that the server solves a standard ML problem to learn its model parameters, i.e., it does not make any attempt to account for fairness or privacy. The fact that it might ``automatically" compute a fair and privacy-preserving model is a direct consequence of how each client computes its synthetic dataset in our framework.

\subsection{Complete Algorithm}
\label{sec:complete_alg}

Instead of choosing nonnegative parameters $\epsilon^{SP}$ and $\epsilon^{EO}$ and solving the \emph{constrained} optimization problem in (\ref{prob:main.orig.client}), we choose to solve a penalty reformulation of (\ref{prob:main.orig.client}) that takes the form of an \emph{unconstrained} optimization problem, for the purposes of computational efficiency and convenience.  Essentially, our penalty problem described below computes a solution to (\ref{prob:main.orig.client}) for \emph{implicitly} defined values of $\epsilon^{SP}$ or $\epsilon^{EO}$.  

For a given choice of penalty parameter $\rho_o \in (0,\infty)$, regularization parameters $(\lambda_{\xhat}, \lambda_\theta) \in (0,\infty)^2$, and previously computed values for $\{(\shat_i,\yhat_i)\}$, we define 
\begin{align*}
\text{(SP measure)} \ \
&\Pcal_{SP} \left(\{\xhat_i\} \right) := \ell \left(\{d_i\},\theta(\{\dhat_i\})\right)
+
\frac{\rho_o}{2N^2}\left(\sum_{i=1}^{N}(s_i-\bar{s})a_i^T \theta(\{\dhat_i\}) \right)^2 +
\frac{\lambda_{\xhat}}{2(N_s^1 \cdot n )^2} \sum_{i=1}^{N_s^1} \|\xhat_i\|_2^2 \ \ \text{and} \\
\text{(EO measure)} \ \ 
&\Pcal_{EO} \left(\{\xhat_i\} \right) := \ell \left(\{d_i\},\theta(\{\dhat_i\})\right)
+
\frac{\rho_o}{2N^2}\left(\sum_{i=1}^{N}(s_i-\bar{s}) \left(\frac{1+y_i}{2}\right) a_i^T \theta(\{\dhat_i\}) \right)^2 +
\frac{\lambda_{\xhat}}{2(N_s^1 \cdot n )^2} \sum_{i=1}^{N_s^1} \|\xhat_i\|_2^2
\end{align*}
with
\begin{equation}
\begin{aligned} \label{prob:reg.inner}
\theta(\{\dhat_i\})
:= \arg\min_{\theta} \ \Biggl\{ \ell_r(\{\dhat_i\},\theta) &:= \ell(\{\dhat_i\},\theta)  + \frac{\lambda_\theta}{2n^2} \|\theta\|_2^2 \Biggr\}.
\end{aligned}
\end{equation}
Then, given our fairness preferences, we can choose to solve any of the following penalty reformulations of~(\ref{prob:main.orig.client}):
\begin{equation} \label{prob:penalty.outer}
\begin{aligned}
\min_{\{\xhat_i\}} \ \Pcal \left(\{\xhat_i\} \right)
\ \ \text{with either} \ \ 
\Pcal = \Pcal_{SP} \ \text{or} \ 
\Pcal = \Pcal_{EO} \ \text{or} \ 
\Pcal = \Pcal_{SP} + \Pcal_{EO}.
\end{aligned}
\end{equation}
(Other fairness measures could  also be incorporated in a straightforward manner.) Here,  $\rho_o$ is the penalty parameter associated with fairness constraints on the client's original training data, and $\lambda_{\xhat}$ and $\lambda_{\theta}$ are  regularization parameters for the outer and inner problems, respectively.  Regularization terms are frequently adopted in ML modeling formulations for a variety of reasons. The regularization used on the nonsensitive features $\{\xhat_i\}$ in the outer problem helps prevent the computation of excessively large values for $\{\xhat_i\}$, whereas the regularization term in the inner problem is introduced primarily to guarantee that the inner problem has a unique minimizer.

When required to solve the inner optimization problem (i.e., problem~\eqref{prob:reg.inner}), any solver designed for unconstrained optimization is applicable.  In our testing, we choose the L-BFGS method with a strong Wolfe line search procedure~\cite{wright2006numerical}.

To solve the outer optimization problem~\eqref{prob:penalty.outer}, we use a descent algorithm.  Computing the gradient is nontrivial because of the inner optimization problem.  That said, the gradient can be computed using the chain rule.  Observe that 

\begin{equation}
\begin{aligned}\label{eqa:grad.P.xhat}
\Pcal'\big( \{\xhat_i\} \big)
= \frac{\partial \Pcal(\{\xhat_i\})}{\partial \theta(\{\dhat_i\})}
\cdot
\frac{\partial \theta(\{\dhat_i\})}{\partial \{\xhat_i\}} + \frac{\lambda_{\xhat}}{(N_s^1 \cdot n )^2} M(\{\xhat_i\})
\end{aligned}
\end{equation}
where $M(\{\xhat_i\})$ has $i$th row $\xhat_i^T$.  Then, $\nabla \Pcal(\{\xhat_i\}) = (\Pcal'(\{\xhat_i\}))^T$.  Computing $\partial \Pcal(\{\xhat_i\}) / \partial \theta(\{\dhat_i\})$ in~\eqref{eqa:grad.P.xhat} is straightforward while computing $\partial \theta(\{\dhat_i\}) / \partial \{\xhat_i\}$ requires implicit differentiation to obtain
\begin{equation}\label{eqa:dtheta.dxhat}
\begin{aligned}
\frac{\partial^2 \ell_r(\{\dhat_i\},\theta(\{\dhat_i\}))}{\partial^2\theta} \cdot
\frac{\partial \theta(\{\dhat_i\})}{\partial \{\xhat_i\}}
=
\frac{\partial^2 \ell_r(\{\dhat_i\},\theta(\{\dhat_i\}))}{\partial\theta\partial \{\xhat_i\}},
\end{aligned}
\end{equation}
which amounts to solving a linear system of equations.  (Recall that $\ell_r$ is defined in~\eqref{prob:reg.inner}.)

With respect to the training performed on the server, we also include a regularization term to be consistent with the inner subproblem~\eqref{prob:reg.inner}.  Specifically, by letting $\dhat_{i, c}$ denote the $i$th synthetic data point computed by client $c$ so that $\{\dhat_{i, c}\}$ denotes the combined synthetic dataset from all clients, the server model parameters are trained by solving
\begin{equation}
\begin{aligned}\label{prob:reg.server}
\theta(\{\dhat_{i, c}\}) := \arg\min_{\theta} \ \ell_r(\{\dhat_{i, c}\},\theta).
\end{aligned}
\end{equation}

After completing the first stage, the synthetic dataset obtained by solving the penalty reformulation~(\ref{prob:penalty.outer}) is used as input for the second stage. In the second stage, clients can utilize any privacy-preserving synthetic data generation method, such as those mentioned in  \cite{jordon2018pate, xie2018differentially, fang2022dp}, so that the final synthetic dataset not only mitigates unfairness but also satisfies certain privacy guarantees, which makes it appropriate for the server-side global model training.

The entire algorithm is shown in Algorithm~\ref{alg:main}. Client training starts at line~\ref{alg.line:client.start}, and server training begins at line~\ref{alg.line:server.start}. Unlike conventional privacy-preserving FL algorithms in the distributed environment, which frequently communicate model parameters back and forth between the server and clients, our CML approach communicates a single time (i.e., passes the synthetic data from clients to server and model parameters from server to client a single time). To quantify the communication cost, let $K$ denote the number of clients, $t$ the number of communication rounds, $|\theta|$ the size of the server's model parameters, and $|D|$ the size of the synthetic dataset for each client (assumed here to be the same for simplicity). For traditional FL, the total communication cost can be expressed as $\mathcal{O}(K \cdot t \cdot |\theta|)$. In contrast, our one-time CML approach has a communication cost of $\mathcal{O}(K \cdot |D|)$ from the clients to the server and a cost of $\mathcal{O}(K \cdot |\theta|)$ from the server back to the clients.  Therefore,  our approach is significantly more efficient in terms of communication cost when $t \cdot |\theta|$ is large relative to $|D| + |\theta|$, which can often be the case since $t$ is often large and $|D|$, which is a user-defined parameter, can be chosen to be relatively small in our approach.

\IncMargin{1em}
\begin{algorithm}[ht]
\caption{CML Method for Ensuring Fairness and Preserving Privacy}
\label{alg:main}




\textbf{Client's Role:} Each client does the following steps. \label{alg.line:client.start} \\
Secure real training dataset $\{d_i\} = \{(x_i, s_i, y_i)\}_{i=1}^{N}$\\
Choose desired synthetic data sizes $N_s^1$ and $N_s^2$ for the first stage and second stage, respectively\\
Choose $\{(\shat_i, \yhat_i)\}$ (see the discussion in Section~\ref{sec.client-side})\\ 
Initialize synthetic data $\{\xhat_{i, 0}\}$ as desired\\
Choose parameters $(\rho_o, \lambda_{\theta}, \lambda_{\xhat}) \in (0,\infty)^3$\\
Choose differential privacy parameters $(\epsilon, \delta) \in (0,\infty)^2$\\
Choose maximum number of iterations $k_{\max}$\\
\textbf{\textit{First-stage: Learning unfairness-mitigated synthetic data}}\\
\For{$k \gets 1$ \KwTo $k_{\max}$}{
    Compute $\theta(\{\dhat_{i, k}\})$ by solving problem (11)\\
    Compute $\frac{\partial \theta(\{\dhat_{i, k}\})}{\partial \{\xhat_{i, k}\}}$ by solving the linear system (14)\\
    Compute $\frac{\partial \mathcal{P}(\{\xhat_{i, k}\})}{\partial \theta(\{\dhat_{i, k}\})}$\\
    Compute $\mathcal{P}'(\{\xhat_{i, k}\})$ as in (13)\\
    $\nabla \mathcal{P}(\{\xhat_{i, k}\}) \gets (\mathcal{P}'(\{\xhat_{i, k}\}))^T$\\
    Update $\{\xhat_{i, k+1}\}$ using $\{\xhat_{i, k}\}$ and $\nabla \Pcal(\{\xhat_{i, k}\})$ (e.g., using gradient descent or Adam solver~\cite{kingma2014adam}) \label{step.update} \\
}
Set $\{\dhat_i\} \gets \{(\xhat_i, \shat_i, \yhat_i)\}_{i=1}^{N_s^1}$ for the second-stage\\
\textbf{\textit{Second-stage: Learning privacy-preserved synthetic data}}\\
Use $\{\dhat_i\}, \epsilon, \delta$, and $N_s^2$ as input to DP-CTGAN to compute revised values $\{\dhat_i\} = \{(\xhat_i, \shat_i, \yhat_i)\}_{i=1}^{N_s^2}$ \\
%
Send $\{\dhat_i\} = \{(\xhat_i, \shat_i, \yhat_i)\}_{i=1}^{N_s^2}$ to the server\\
\algrule
\textbf{Server's Role:} \label{alg.line:server.start} \\
Obtain combined client synthetic data $\{\dhat_{i, c}\}$, where $\dhat_{i,c}$ denotes the synthetic dataset sent by client $c$ \\
Train server model by computing $\theta(\{\dhat_{i, c}\})$ from (15) \\
Communicate $\theta(\{\dhat_{i, c}\})$ to the clients
\end{algorithm}
\DecMargin{1em}

\section{Experiments} \label{sec:experiments}
In this section, we test through numerical experiments the performance of our approach. In particular, we examine how the penalty parameters and the size of the synthetic data affect the trade-off between accuracy and fairness metrics.

\subsection{Experimental Setup}

\textbf{Computing environment.}
The experiments reported in this paper were conducted on a machine equipped with an AMD Ryzen 9 7900X3D 12-Core CPU Processor. The proposed framework was simulated using client parallel training implemented via the Python package \texttt{multiprocessing}.

\smallskip
\noindent
\textbf{Datasets.} 
We present experiments for our approach on the binary classification datasets \texttt{Law School} \cite{wightman1998lsac}, 
which is a real-world dataset widely used in the fairness-aware literature \cite{salazar2023fair}. The \texttt{Law School} dataset contains 20,798 data points and consists of 11 features. We treat the \textit{race} column as our sensitive feature. The goal is to predict whether a student will pass the bar exam. We split the dataset evenly into two clients and randomly allocated 80\% of the data for training and 20\% for testing.
Our tests use a relatively small number of clients for our purpose of demonstration.  We expect that our results are representative of other settings as well (i.e., when $K \gg 0$).

\smallskip
\noindent
\textbf{Hyperparameters and algorithm choices.}
We set the regularization parameter $\lambda_{\xhat} = 10^{-4}$. The maximum number of L-BFGS iterations used when solving the inner problem~\eqref{prob:reg.inner} is $100$, and $k_{max} = 1000$ in Algorithm~\ref{alg:main}.  The iterate update in Line~\ref{step.update} of Algorithm~\ref{alg:main} is performed using \textit{Adam} \cite{kingma2014adam}.  Other hyperparameters for the outer problem~\eqref{prob:penalty.outer} include the penalty parameter $\rho_o \in \{0, 10, 100, 1000, 10000\}$, the regularization parameter $\lambda_\theta=10^{-4}$, the first-stage synthetic dataset size $N_s^1$ being equivalent to the real training dataset size $N$, and the second-stage synthetic data size $N_S^2$ being either $N$ or $0.1N$.

\subsection{Experimental Results}

\begin{figure*}[t]
\centering
\includegraphics[width=0.99\textwidth]{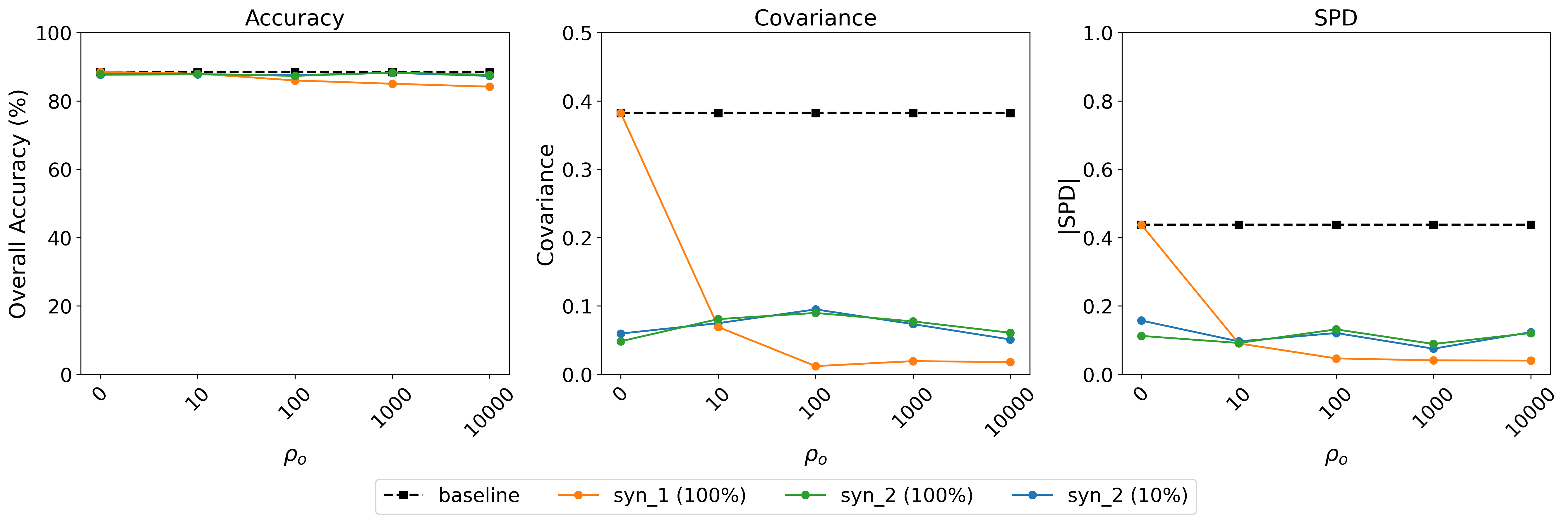} 
\caption{Performance trends of the accuracy, the covariance estimate (see~\eqref{equ:DBC}), and the absolute value of the \SPD{} fairness measure (see~\eqref{equ:fairness-measures}) for different values of the penalty parameter $\rho_o$,  computed using the client testing data on the \texttt{Law School} dataset. Each plot compares the following: the baseline corresponds to $\rho_0 = 0$ so that fairness is completely ignored,  syn\textunderscore1(100\%) is the first-stage synthetic dataset with $N_s^1 = N$, syn\textunderscore2(100\%) is the second-stage synthetic dataset with $N_s^2 = N$, and syn\textunderscore2(10\%) is the second-stage synthetic dataset with $N_s^2 = 0.1N$.}
\label{fig:Law_SP}
\Description{This figure presents four line charts showing the trends of accuracy, covariance, correlation, and \SPD{} fairness metrics as the penalty parameter $\rho_o$ increases. The data points represent performance measured on the \texttt{Law School} dataset. The four charts compare the real dataset, and the synthetic dataset in different stages of different sizes. The black dashed line corresponds to the real dataset, while the orange, green, and blue lines represent synthetic datasets (\texttt{syn_1(100\%)}, \texttt{syn_2(100\%)}, and \texttt{syn_2(10\%)}, respectively). Each metric illustrates the trade-off between accuracy and fairness as $\rho_o$ increases. With covariance and fairness measures (\SPD{}) approaching zero for larger $\rho_o$, indicates reduced unfairness. The synthetic datasets show varying performance trends, reflecting their stage and size differences relative to the real dataset.}
\end{figure*}

\begin{figure*}[t]
\centering
\includegraphics[width=0.99\textwidth]{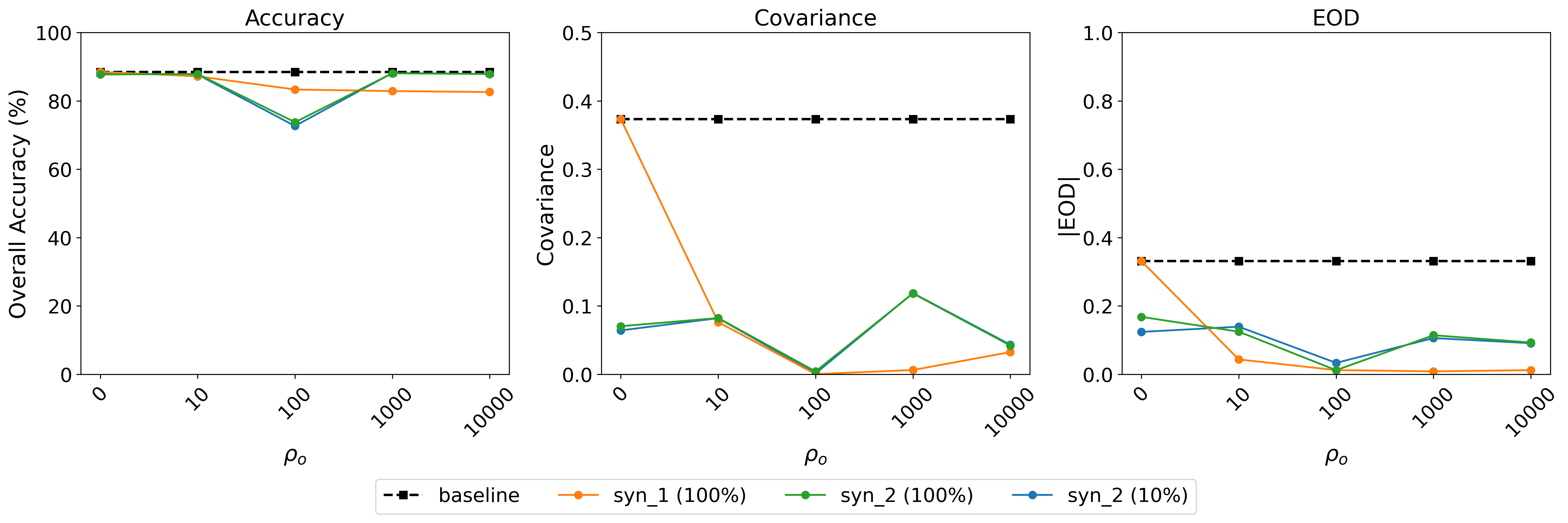} 
\caption{Performance trends of the accuracy, the covariance estimate (see~\eqref{equ:DBC}), and the absolute value of the \EOD{} fairness measure (see~\eqref{equ:fairness-measures}) for different values of the penalty parameter $\rho_o$, computed using the client testing data on the \texttt{Law School} dataset. Each plot compares the following: the baseline corresponds to $\rho_0$ so that fairness is completely ignored,  syn\textunderscore1(100\%) is the first-stage synthetic dataset with $N_s^1 = N$, syn\textunderscore2(100\%) is the second-stage synthetic dataset with $N_s^2 = N$, and syn\textunderscore2(10\%) is the second-stage synthetic dataset with $N_s^2 = 0.1N$.}
\label{fig:Law_EO}
\Description{This figure presents four line charts showing the trends of accuracy, covariance, correlation, and \EOD{} fairness metrics as the penalty parameter $\rho_o$ increases. The data points represent performance measured on the \texttt{Law School} dataset. The four charts compare the real dataset, and the synthetic dataset in different stages of different sizes. The black dashed line corresponds to the real dataset, while the orange, green, and blue lines represent synthetic datasets (\texttt{syn_1(100\%)}, \texttt{syn_2(100\%)}, and \texttt{syn_2(10\%)}, respectively). Each metric illustrates the trade-off between accuracy and fairness as $\rho_o$ increases. With covariance and fairness measures (\EOD{}) approaching zero for larger $\rho_o$, indicates reduced unfairness. The synthetic datasets show varying performance trends, reflecting their stage and size differences relative to the real dataset.}
\end{figure*}

\textbf{Ability to Mitigate Unfairness.} 
To assess the ability of our method to mitigate unfairness, we first examine the trade-off between the first stage synthetic dataset's accuracy lost and fairness gained by choosing different values for the penalty parameters  $\rho_o$, which correspond to the fairness constraints associated with the client's original training data. For each client, we set $N_s^1 = N$ and $\{(\shat_i, \yhat_i)\} = \{(s_i, y_i)\}$, and initialize $\{\xhat_{i, 0}\} = \{x_i\}$, so that we may clearly observe whether the first-stage synthetic datasets have the ability to mitigate unfairness. We conduct our tests on the \texttt{Law School} dataset, where race is the sensitive feature.  The results are depicted in Figure~\ref{fig:Law_SP} and Figure~\ref{fig:Law_EO} for fairness measures SP and EO, respectively. The values presented in the figures are computed across all clients' testing data.

The black dashed lines in Figure~\ref{fig:Law_SP} and Figure~\ref{fig:Law_EO} are the  baselines since they correspond to $\rho_o = 0$, i.e., no penalization on unfairness is enforced. In this case, the first stage synthetic data $\{\dhat_i\} = \{d_i\}$, which means that the solution to the outer problem~\eqref{prob:penalty.outer} is identical to the initialization of $\{\xhat_i\}$, i.e., $\{x_i\}$. This scenario is equivalent to evaluating fairness metrics on the client's real dataset. Note that the orange lines in these figures (i.e.,  \texttt{syn\textunderscore1(100\%)}) illustrate the performance trends of the first-stage unfairness-mitigated synthetic data, which is equal in size to the client's real data. As $\rho_o$ increases, we can see that the accuracy generally decreases slightly. The covariance estimates, which we aim to reduce via the penalty term when solving the outer optimization problem~\eqref{prob:penalty.outer}, decrease significantly compared to the baselines and show a clear trend toward zero. 
Furthermore, the absolute values of \SPD{} and \EOD{} measures (i.e., |\SPD{}| and |\EOD{}|) approach zero, thus reflecting a decrease in unfairness with respect to the group fairness measures SP and EO. 

The specific values for the accuracy and fairness metrics are presented in Table~\ref{tbl:law_SPD} and Table~\ref{tbl:law_EOD}. Based on these results, one might be tempted to choose $\rho_o = 10$, which leads to a mere $0.51\%$ drop in accuracy and a significant improvement in |\SPD{}| from 0.4377 to 0.0903, or to sacrifice $1.28\%$ in accuracy to improve |\EOD{}| from 0.3313 to 0.0435. 

\begin{table*}[t]
\caption{Accuracy and |SPD| performance measures for various $\rho_o$ values on the \texttt{Law School} dataset. Synthetic Data 1 (resp., Synthetic Data 2) refers to first-stage (resp., second-stage) synthetic data. Values with an asterisk refer to the baseline.
}
\label{tbl:law_SPD}
\centering
\begin{tabular}{cccccccc}
    \toprule
     & \multicolumn{2}{c}{Synthetic Data 1 (100\%)} & \multicolumn{2}{c}{Synthetic Data 2 (100\%)} & \multicolumn{2}{c}{Synthetic Data 2 (10\%)} \\
     \cmidrule(lr){2-3} \cmidrule(lr){4-5} \cmidrule(lr){6-7}
    $\rho_o$ & Accuracy (\%) & |SPD| & Accuracy (\%) & |SPD| & Accuracy (\%) & |SPD| \\
    \midrule
    0     & 88.49* & 0.4377* & 87.88 & 0.1123 & 87.67 & 0.1572 \\
    10    & 87.98 & 0.0903 & 87.86 & 0.0917 & 87.79 & 0.0963 \\
    100   & 85.99 & 0.0465 & 87.33 & 0.1315 & 87.55 & 0.1208 \\
    1000  & 85.02 & 0.0408 & 88.29 & \textbf{0.0887} & 88.22 & \textbf{0.0749} \\
    10000 & 84.18 & 0.0401 & 87.64 & 0.1208 & 87.31 & 0.1234 \\
    \bottomrule
\end{tabular}
\end{table*}

\begin{table*}[t]
\caption{Accuracy and |EOD| performance measures for various $\rho_o$ values on the \texttt{Law School} dataset. Synthetic Data 1 (resp., Synthetic Data~2) refers to first-stage (resp., second-stage) synthetic data. Values with an asterisk refer to the baseline.
}
\label{tbl:law_EOD}
\centering
\begin{tabular}{cccccccc}
    \toprule
     & \multicolumn{2}{c}{Synthetic Data 1 (100\%)} & \multicolumn{2}{c}{Synthetic Data 2 (100\%)} & \multicolumn{2}{c}{Synthetic Data 2 (10\%)} \\
     \cmidrule(lr){2-3} \cmidrule(lr){4-5} \cmidrule(lr){6-7}
    $\rho_o$ & Accuracy (\%) & |EOD| & Accuracy (\%) & |EOD| & Accuracy (\%) & |EOD| \\
    \midrule
    0     & 88.49* & 0.3313* & 87.74 & 0.1681 & 88.08 & 0.1241 \\
    10    & 87.21 & 0.0435 & 87.88 & 0.1247 & 87.72 & 0.1393 \\
    100   & 83.34 & 0.0122 & 73.77 & 0.0120 & 72.67 & 0.0334 \\
    1000  & 82.88 & 0.0086 & 88.08 & 0.1143 & 88.17 & 0.1060 \\
    10000 & 82.60 & 0.0121 & 87.88 & \textbf{0.0931} & 87.91 & \textbf{0.0910} \\
    \bottomrule
\end{tabular}
\end{table*}

\smallskip
\noindent
\textbf{Ability to Preserve Privacy.}
We begin by recalling that in the second stage of our pipeline, we apply a data generation method with differential-privacy guarantees (i.e., DP-CTGAN \cite{fang2022dp}) with our first-stage unfairness-mitigated synthetic data as input. In DP-CTGAN, random noise is added to the discriminator's gradients during the training process so that the synthetic data generated from the generator adheres to $(\epsilon, \delta)$-differential privacy \cite{dwork2014algorithmic}. In our experiments, we use the default settings of DP-CTGAN where $\epsilon=3$ and $\delta=10^{-5}$. This implies that, even in the (generally unlikely) case where the solution of problem~\eqref{prob:penalty.outer} results in $\{\dhat_i\} = \{d_i\}$, which means that the input to the second stage is identical to the client's real dataset, the output still satisfies $(\epsilon, \delta)$-differential privacy. Consequently, it is very difficult to determine if a single data point exists in this dataset or not, which aids in ensuring privacy preservation.

In Figure~\ref{fig:Law_SP} and Figure~\ref{fig:Law_EO}, the green lines (i.e., the ones corresponding to  \texttt{syn\textunderscore2 (100\%)}) represent the second stage synthetic dataset output from DP-CTGAN of size $N_s^2 = N$ (computed using the first-stage synthetic data  of size $N_s^1 = N$ as input). Except for $\rho_o=100$ in the \EOD{} metric, the accuracy is within 1\% of the baseline. As for the covariances, although they are not monotonically decreasing, they are still smaller than the baseline. 

From the detailed results shown in Table~\ref{tbl:law_SPD} and Table~\ref{tbl:law_EOD}, one can select $\rho_o = 1000$ as it results in only a $0.2\%$ loss of accuracy while significantly improving |\SPD{}| from 0.4377 to 0.0887. Alternatively, choosing $\rho_o=10000$ is also appealing in terms of the \EOD{} metric since it sacrifices $0.61\%$ in accuracy to achieve an improvement in |\EOD{}| from 0.3313 to 0.0931.

\smallskip
\noindent
\textbf{Ability to Reduce Communication Costs.} 
We can control the communication costs of our method by controlling the sizes of the synthetic datasets that are passed to the server.  Recall that $N_s^1$ and $N_s^2$ are the desired synthetic data sizes chosen by the user for the first and second stages, respectively. We conducted experiments with $N_s^2 \ll N_s^1 = N$ in the second stage. The blue lines in Figure~\ref{fig:Law_SP} and Figure~\ref{fig:Law_EO} (i.e.,  \texttt{syn\textunderscore2 (10\%)}) are the results using the second-stage synthetic dataset output by DP-CTGAN with data size $N_s^2 = 0.1N$ and the first-stage synthetic dataset  of size $N_s^1 = N$ as input. The trends for each metric are almost the same as for  \texttt{syn\textunderscore2 (100\%)}, indicating that instead of transferring a  second-stage synthetic data of the same size as the real data size to the server, we can transfer a data set of only 10\% of the size and achieve nearly the same performance.

Consider the results in Table~\ref{tbl:law_SPD} and Table~\ref{tbl:law_EOD} with, e.g., $\rho_o = 1000$.  If one passes a synthetic data set of size $N_s^2 = N$, then |\SPD{}| improves from the baseline of $0.4377$ to $0.0887$, whereas if one passes a synthetic data set of size $N_s^2 = 0.1N$, then |\SPD{}| improves from the baseline of $0.4377$ to $0.0749$.  At the same time, with $N_s^2 = N$, there is a 0.2\% loss in accuracy, while with $N_s^2 = 0.1N$ the loss in accuracy is a comparable 0.27\%. The results with respect to |\EOD{}| again show comparable performance whether $N_s^2 = N$ or $N_s^2 = 0.1N$.

\section{Conclusion}
In this study, we present a two-stage cost-effective communication strategy to mitigate unfairness and help preserve privacy across clients in a CML context. We propose to send a synthetic dataset from each client to the server instead of each client's model, as is often done in a distributed environment. The manner in which we propose computing the synthetic dataset takes unfairness into account by solving a bilevel optimization problem in the first stage, and utilizing a differential privacy-guaranteeing synthetic data generation method to obtain a second data set in the second stage. This allows the server to train a conventional ML model without further fairness or privacy concerns on the collection of client-provided synthetic datasets. This approach not only simplifies the training process for the server but also significantly reduces the communication cost by limiting it to a one-time transfer between the clients and the the server. Our implementation on a well-known dataset illustrates that our approach effectively mitigates unfairness and preserves client privacy without sacrificing too much accuracy. Even with synthetic datasets of 10\% the size of the client training data, our strategy remains effective in producing accurate models that are fair and aid in maintaining privacy.

\bibliographystyle{ACM-Reference-Format}
\bibliography{sample-base}


\begin{thebibliography}{38}


\ifx \showCODEN    \undefined \def \showCODEN     #1{\unskip}     \fi
\ifx \showDOI      \undefined \def \showDOI       #1{#1}\fi
\ifx \showISBNx    \undefined \def \showISBNx     #1{\unskip}     \fi
\ifx \showISBNxiii \undefined \def \showISBNxiii  #1{\unskip}     \fi
\ifx \showISSN     \undefined \def \showISSN      #1{\unskip}     \fi
\ifx \showLCCN     \undefined \def \showLCCN      #1{\unskip}     \fi
\ifx \shownote     \undefined \def \shownote      #1{#1}          \fi
\ifx \showarticletitle \undefined \def \showarticletitle #1{#1}   \fi
\ifx \showURL      \undefined \def \showURL       {\relax}        \fi
\providecommand\bibfield[2]{#2}
\providecommand\bibinfo[2]{#2}
\providecommand\natexlab[1]{#1}
\providecommand\showeprint[2][]{arXiv:#2}

\bibitem[Abadi et~al\mbox{.}(2016)]%
        {abadi2016deep}
\bibfield{author}{\bibinfo{person}{Martin Abadi}, \bibinfo{person}{Andy Chu}, \bibinfo{person}{Ian Goodfellow}, \bibinfo{person}{H~Brendan McMahan}, \bibinfo{person}{Ilya Mironov}, \bibinfo{person}{Kunal Talwar}, {and} \bibinfo{person}{Li Zhang}.} \bibinfo{year}{2016}\natexlab{}.
\newblock \showarticletitle{Deep learning with differential privacy}. In \bibinfo{booktitle}{\emph{Proceedings of the 2016 ACM SIGSAC conference on computer and communications security}}. \bibinfo{pages}{308--318}.
\newblock


\bibitem[{Brendan McMahan and Daniel Ramage}(2017)]%
        {googleblog}
\bibfield{author}{\bibinfo{person}{{Brendan McMahan and Daniel Ramage}}.} \bibinfo{year}{2017}\natexlab{}.
\newblock \bibinfo{title}{Google AI blog: Federated Learning: Collaborative Machine Learning without Centralized Training Data}.
\newblock
\newblock
\urldef\tempurl%
\url{https://research.google/blog/federated-learning-collaborative-machine-learning-without-centralized-training-data/}
\showURL{%
Retrieved January 12, 2025 from \tempurl}


\bibitem[Buolamwini and Gebru(2018)]%
        {buolamwini2018gender}
\bibfield{author}{\bibinfo{person}{Joy Buolamwini} {and} \bibinfo{person}{Timnit Gebru}.} \bibinfo{year}{2018}\natexlab{}.
\newblock \showarticletitle{Gender shades: Intersectional accuracy disparities in commercial gender classification}. In \bibinfo{booktitle}{\emph{Conference on fairness, accountability and transparency}}. PMLR, \bibinfo{pages}{77--91}.
\newblock


\bibitem[Caton and Haas(2024)]%
        {caton2024fairness}
\bibfield{author}{\bibinfo{person}{Simon Caton} {and} \bibinfo{person}{Christian Haas}.} \bibinfo{year}{2024}\natexlab{}.
\newblock \showarticletitle{Fairness in machine learning: A survey}.
\newblock \bibinfo{journal}{\emph{Comput. Surveys}} \bibinfo{volume}{56}, \bibinfo{number}{7} (\bibinfo{year}{2024}), \bibinfo{pages}{1--38}.
\newblock


\bibitem[d'Alessandro et~al\mbox{.}(2017)]%
        {d2017conscientious}
\bibfield{author}{\bibinfo{person}{Brian d'Alessandro}, \bibinfo{person}{Cathy O'Neil}, {and} \bibinfo{person}{Tom LaGatta}.} \bibinfo{year}{2017}\natexlab{}.
\newblock \showarticletitle{Conscientious classification: A data scientist's guide to discrimination-aware classification}.
\newblock \bibinfo{journal}{\emph{Big data}} \bibinfo{volume}{5}, \bibinfo{number}{2} (\bibinfo{year}{2017}), \bibinfo{pages}{120--134}.
\newblock


\bibitem[Dressel and Farid(2018)]%
        {dressel2018accuracy}
\bibfield{author}{\bibinfo{person}{Julia Dressel} {and} \bibinfo{person}{Hany Farid}.} \bibinfo{year}{2018}\natexlab{}.
\newblock \showarticletitle{The accuracy, fairness, and limits of predicting recidivism}.
\newblock \bibinfo{journal}{\emph{Science advances}} \bibinfo{volume}{4}, \bibinfo{number}{1} (\bibinfo{year}{2018}), \bibinfo{pages}{eaao5580}.
\newblock


\bibitem[Dwork(2006)]%
        {dwork2006differential}
\bibfield{author}{\bibinfo{person}{Cynthia Dwork}.} \bibinfo{year}{2006}\natexlab{}.
\newblock \showarticletitle{Differential privacy}. In \bibinfo{booktitle}{\emph{International colloquium on automata, languages, and programming}}. Springer, \bibinfo{pages}{1--12}.
\newblock


\bibitem[Dwork et~al\mbox{.}(2014)]%
        {dwork2014algorithmic}
\bibfield{author}{\bibinfo{person}{Cynthia Dwork}, \bibinfo{person}{Aaron Roth}, {et~al\mbox{.}}} \bibinfo{year}{2014}\natexlab{}.
\newblock \showarticletitle{The algorithmic foundations of differential privacy}.
\newblock \bibinfo{journal}{\emph{Foundations and Trends{\textregistered} in Theoretical Computer Science}} \bibinfo{volume}{9}, \bibinfo{number}{3--4} (\bibinfo{year}{2014}), \bibinfo{pages}{211--407}.
\newblock


\bibitem[Ezzeldin et~al\mbox{.}(2023)]%
        {ezzeldin2023fairfed}
\bibfield{author}{\bibinfo{person}{Yahya~H Ezzeldin}, \bibinfo{person}{Shen Yan}, \bibinfo{person}{Chaoyang He}, \bibinfo{person}{Emilio Ferrara}, {and} \bibinfo{person}{A~Salman Avestimehr}.} \bibinfo{year}{2023}\natexlab{}.
\newblock \showarticletitle{Fairfed: Enabling group fairness in federated learning}. In \bibinfo{booktitle}{\emph{Proceedings of the AAAI conference on artificial intelligence}}, Vol.~\bibinfo{volume}{37}. \bibinfo{pages}{7494--7502}.
\newblock


\bibitem[Fang et~al\mbox{.}(2022)]%
        {fang2022dp}
\bibfield{author}{\bibinfo{person}{Mei~Ling Fang}, \bibinfo{person}{Devendra~Singh Dhami}, {and} \bibinfo{person}{Kristian Kersting}.} \bibinfo{year}{2022}\natexlab{}.
\newblock \showarticletitle{Dp-ctgan: Differentially private medical data generation using CTGANs}. In \bibinfo{booktitle}{\emph{International Conference on Artificial Intelligence in Medicine}}. Springer, \bibinfo{pages}{178--188}.
\newblock


\bibitem[{Flaticon}(2023)]%
        {flaticon}
\bibfield{author}{\bibinfo{person}{{Flaticon}}.} \bibinfo{year}{2023}\natexlab{}.
\newblock \bibinfo{title}{Illustration Icons}.
\newblock
\newblock
\urldef\tempurl%
\url{https://www.flaticon.com/}
\showURL{%
Retrieved June 02, 2023 from \tempurl}


\bibitem[Giovannelli et~al\mbox{.}(2024)]%
        {giovannelli2024bilevel}
\bibfield{author}{\bibinfo{person}{Tommaso Giovannelli}, \bibinfo{person}{Griffin~Dean Kent}, {and} \bibinfo{person}{Luis~Nunes Vicente}.} \bibinfo{year}{2024}\natexlab{}.
\newblock \showarticletitle{Bilevel optimization with a multi-objective lower-level problem: Risk-neutral and risk-averse formulations}.
\newblock \bibinfo{journal}{\emph{Optimization Methods and Software}} (\bibinfo{year}{2024}), \bibinfo{pages}{1--23}.
\newblock


\bibitem[Goetz and Tewari(2020)]%
        {goetz2020federated}
\bibfield{author}{\bibinfo{person}{Jack Goetz} {and} \bibinfo{person}{Ambuj Tewari}.} \bibinfo{year}{2020}\natexlab{}.
\newblock \showarticletitle{Federated learning via synthetic data}.
\newblock \bibinfo{journal}{\emph{arXiv preprint arXiv:2008.04489}} (\bibinfo{year}{2020}).
\newblock


\bibitem[Goodfellow et~al\mbox{.}(2020)]%
        {goodfellow2020generative}
\bibfield{author}{\bibinfo{person}{Ian Goodfellow}, \bibinfo{person}{Jean Pouget-Abadie}, \bibinfo{person}{Mehdi Mirza}, \bibinfo{person}{Bing Xu}, \bibinfo{person}{David Warde-Farley}, \bibinfo{person}{Sherjil Ozair}, \bibinfo{person}{Aaron Courville}, {and} \bibinfo{person}{Yoshua Bengio}.} \bibinfo{year}{2020}\natexlab{}.
\newblock \showarticletitle{Generative adversarial networks}.
\newblock \bibinfo{journal}{\emph{Commun. ACM}} \bibinfo{volume}{63}, \bibinfo{number}{11} (\bibinfo{year}{2020}), \bibinfo{pages}{139--144}.
\newblock


\bibitem[Hu et~al\mbox{.}(2022)]%
        {hu2022fedsynth}
\bibfield{author}{\bibinfo{person}{Shengyuan Hu}, \bibinfo{person}{Jack Goetz}, \bibinfo{person}{Kshitiz Malik}, \bibinfo{person}{Hongyuan Zhan}, \bibinfo{person}{Zhe Liu}, {and} \bibinfo{person}{Yue Liu}.} \bibinfo{year}{2022}\natexlab{}.
\newblock \showarticletitle{Fedsynth: Gradient compression via synthetic data in federated learning}.
\newblock \bibinfo{journal}{\emph{arXiv preprint arXiv:2204.01273}} (\bibinfo{year}{2022}).
\newblock


\bibitem[Jordon et~al\mbox{.}(2018)]%
        {jordon2018pate}
\bibfield{author}{\bibinfo{person}{James Jordon}, \bibinfo{person}{Jinsung Yoon}, {and} \bibinfo{person}{Mihaela Van Der~Schaar}.} \bibinfo{year}{2018}\natexlab{}.
\newblock \showarticletitle{PATE-GAN: Generating synthetic data with differential privacy guarantees}. In \bibinfo{booktitle}{\emph{International conference on learning representations}}.
\newblock


\bibitem[Kim et~al\mbox{.}(2021)]%
        {kim2021age}
\bibfield{author}{\bibinfo{person}{Eugenia Kim}, \bibinfo{person}{De'Aira Bryant}, \bibinfo{person}{Deepak Srikanth}, {and} \bibinfo{person}{Ayanna Howard}.} \bibinfo{year}{2021}\natexlab{}.
\newblock \showarticletitle{Age bias in emotion detection: An analysis of facial emotion recognition performance on young, middle-aged, and older adults}. In \bibinfo{booktitle}{\emph{Proceedings of the 2021 AAAI/ACM Conference on AI, Ethics, and Society}}. \bibinfo{pages}{638--644}.
\newblock


\bibitem[Kingma(2014)]%
        {kingma2014adam}
\bibfield{author}{\bibinfo{person}{DP Kingma}.} \bibinfo{year}{2014}\natexlab{}.
\newblock \showarticletitle{Adam: a method for stochastic optimization}.
\newblock \bibinfo{journal}{\emph{arXiv preprint arXiv:1412.6980}} (\bibinfo{year}{2014}).
\newblock


\bibitem[Laal and Laal(2012)]%
        {laal2012collaborative}
\bibfield{author}{\bibinfo{person}{Marjan Laal} {and} \bibinfo{person}{Mozhgan Laal}.} \bibinfo{year}{2012}\natexlab{}.
\newblock \showarticletitle{Collaborative learning: what is it?}
\newblock \bibinfo{journal}{\emph{Procedia-Social and Behavioral Sciences}}  \bibinfo{volume}{31} (\bibinfo{year}{2012}), \bibinfo{pages}{491--495}.
\newblock


\bibitem[Lohia et~al\mbox{.}(2019)]%
        {lohia2019bias}
\bibfield{author}{\bibinfo{person}{Pranay~K Lohia}, \bibinfo{person}{Karthikeyan~Natesan Ramamurthy}, \bibinfo{person}{Manish Bhide}, \bibinfo{person}{Diptikalyan Saha}, \bibinfo{person}{Kush~R Varshney}, {and} \bibinfo{person}{Ruchir Puri}.} \bibinfo{year}{2019}\natexlab{}.
\newblock \showarticletitle{Bias mitigation post-processing for individual and group fairness}. In \bibinfo{booktitle}{\emph{Icassp 2019-2019 ieee international conference on acoustics, speech and signal processing (icassp)}}. IEEE, \bibinfo{pages}{2847--2851}.
\newblock


\bibitem[McMahan et~al\mbox{.}(2017)]%
        {mcmahan2017communication}
\bibfield{author}{\bibinfo{person}{Brendan McMahan}, \bibinfo{person}{Eider Moore}, \bibinfo{person}{Daniel Ramage}, \bibinfo{person}{Seth Hampson}, {and} \bibinfo{person}{Blaise~Aguera y Arcas}.} \bibinfo{year}{2017}\natexlab{}.
\newblock \showarticletitle{Communication-efficient learning of deep networks from decentralized data}. In \bibinfo{booktitle}{\emph{Artificial intelligence and statistics}}. PMLR, \bibinfo{pages}{1273--1282}.
\newblock


\bibitem[Mehrabi et~al\mbox{.}(2021)]%
        {mehrabi2021survey}
\bibfield{author}{\bibinfo{person}{Ninareh Mehrabi}, \bibinfo{person}{Fred Morstatter}, \bibinfo{person}{Nripsuta Saxena}, \bibinfo{person}{Kristina Lerman}, {and} \bibinfo{person}{Aram Galstyan}.} \bibinfo{year}{2021}\natexlab{}.
\newblock \showarticletitle{A survey on bias and fairness in machine learning}.
\newblock \bibinfo{journal}{\emph{ACM computing surveys (CSUR)}} \bibinfo{volume}{54}, \bibinfo{number}{6} (\bibinfo{year}{2021}), \bibinfo{pages}{1--35}.
\newblock


\bibitem[Mugunthan et~al\mbox{.}(2021)]%
        {mugunthan2021bias}
\bibfield{author}{\bibinfo{person}{Vaikkunth Mugunthan}, \bibinfo{person}{Vignesh Gokul}, \bibinfo{person}{Lalana Kagal}, {and} \bibinfo{person}{Shlomo Dubnov}.} \bibinfo{year}{2021}\natexlab{}.
\newblock \showarticletitle{Bias-free fedgan: A federated approach to generate bias-free datasets}.
\newblock \bibinfo{journal}{\emph{arXiv preprint arXiv:2103.09876}} (\bibinfo{year}{2021}).
\newblock


\bibitem[Obermeyer et~al\mbox{.}(2019)]%
        {obermeyer2019dissecting}
\bibfield{author}{\bibinfo{person}{Ziad Obermeyer}, \bibinfo{person}{Brian Powers}, \bibinfo{person}{Christine Vogeli}, {and} \bibinfo{person}{Sendhil Mullainathan}.} \bibinfo{year}{2019}\natexlab{}.
\newblock \showarticletitle{Dissecting racial bias in an algorithm used to manage the health of populations}.
\newblock \bibinfo{journal}{\emph{Science}} \bibinfo{volume}{366}, \bibinfo{number}{6464} (\bibinfo{year}{2019}), \bibinfo{pages}{447--453}.
\newblock


\bibitem[Pan et~al\mbox{.}(2023)]%
        {pan2023fedmdfg}
\bibfield{author}{\bibinfo{person}{Zibin Pan}, \bibinfo{person}{Shuyi Wang}, \bibinfo{person}{Chi Li}, \bibinfo{person}{Haijin Wang}, \bibinfo{person}{Xiaoying Tang}, {and} \bibinfo{person}{Junhua Zhao}.} \bibinfo{year}{2023}\natexlab{}.
\newblock \showarticletitle{Fedmdfg: Federated learning with multi-gradient descent and fair guidance}. In \bibinfo{booktitle}{\emph{Proceedings of the AAAI Conference on Artificial Intelligence}}, Vol.~\bibinfo{volume}{37}. \bibinfo{pages}{9364--9371}.
\newblock


\bibitem[Pessach and Shmueli(2022)]%
        {pessach2022review}
\bibfield{author}{\bibinfo{person}{Dana Pessach} {and} \bibinfo{person}{Erez Shmueli}.} \bibinfo{year}{2022}\natexlab{}.
\newblock \showarticletitle{A review on fairness in machine learning}.
\newblock \bibinfo{journal}{\emph{ACM Computing Surveys (CSUR)}} \bibinfo{volume}{55}, \bibinfo{number}{3} (\bibinfo{year}{2022}), \bibinfo{pages}{1--44}.
\newblock


\bibitem[Rasouli et~al\mbox{.}(2020)]%
        {rasouli2020fedgan}
\bibfield{author}{\bibinfo{person}{Mohammad Rasouli}, \bibinfo{person}{Tao Sun}, {and} \bibinfo{person}{Ram Rajagopal}.} \bibinfo{year}{2020}\natexlab{}.
\newblock \showarticletitle{Fedgan: Federated generative adversarial networks for distributed data}.
\newblock \bibinfo{journal}{\emph{arXiv preprint arXiv:2006.07228}} (\bibinfo{year}{2020}).
\newblock


\bibitem[Salazar et~al\mbox{.}(2023)]%
        {salazar2023fair}
\bibfield{author}{\bibinfo{person}{Teresa Salazar}, \bibinfo{person}{Miguel Fernandes}, \bibinfo{person}{Helder Ara{\'u}jo}, {and} \bibinfo{person}{Pedro~Henriques Abreu}.} \bibinfo{year}{2023}\natexlab{}.
\newblock \showarticletitle{Fair-fate: Fair federated learning with momentum}. In \bibinfo{booktitle}{\emph{International Conference on Computational Science}}. Springer, \bibinfo{pages}{524--538}.
\newblock


\bibitem[Savard and Gauvin(1994)]%
        {savard1994steepest}
\bibfield{author}{\bibinfo{person}{Gilles Savard} {and} \bibinfo{person}{Jacques Gauvin}.} \bibinfo{year}{1994}\natexlab{}.
\newblock \showarticletitle{The steepest descent direction for the nonlinear bilevel programming problem}.
\newblock \bibinfo{journal}{\emph{Operations Research Letters}} \bibinfo{volume}{15}, \bibinfo{number}{5} (\bibinfo{year}{1994}), \bibinfo{pages}{265--272}.
\newblock


\bibitem[Sinha et~al\mbox{.}(2017)]%
        {sinha2017review}
\bibfield{author}{\bibinfo{person}{Ankur Sinha}, \bibinfo{person}{Pekka Malo}, {and} \bibinfo{person}{Kalyanmoy Deb}.} \bibinfo{year}{2017}\natexlab{}.
\newblock \showarticletitle{A review on bilevel optimization: From classical to evolutionary approaches and applications}.
\newblock \bibinfo{journal}{\emph{IEEE transactions on evolutionary computation}} \bibinfo{volume}{22}, \bibinfo{number}{2} (\bibinfo{year}{2017}), \bibinfo{pages}{276--295}.
\newblock


\bibitem[Wang et~al\mbox{.}(2022)]%
        {wang2022collaborative}
\bibfield{author}{\bibinfo{person}{Junbo Wang}, \bibinfo{person}{Amitangshu Pal}, \bibinfo{person}{Qinglin Yang}, \bibinfo{person}{Krishna Kant}, \bibinfo{person}{Kaiming Zhu}, {and} \bibinfo{person}{Song Guo}.} \bibinfo{year}{2022}\natexlab{}.
\newblock \showarticletitle{Collaborative machine learning: Schemes, robustness, and privacy}.
\newblock \bibinfo{journal}{\emph{IEEE Transactions on Neural Networks and Learning Systems}} \bibinfo{volume}{34}, \bibinfo{number}{12} (\bibinfo{year}{2022}), \bibinfo{pages}{9625--9642}.
\newblock


\bibitem[Wang et~al\mbox{.}(2018)]%
        {wang2018dataset}
\bibfield{author}{\bibinfo{person}{Tongzhou Wang}, \bibinfo{person}{Jun-Yan Zhu}, \bibinfo{person}{Antonio Torralba}, {and} \bibinfo{person}{Alexei~A Efros}.} \bibinfo{year}{2018}\natexlab{}.
\newblock \showarticletitle{Dataset distillation}.
\newblock \bibinfo{journal}{\emph{arXiv preprint arXiv:1811.10959}} (\bibinfo{year}{2018}).
\newblock


\bibitem[Wang et~al\mbox{.}(2021)]%
        {wang2021enhancing}
\bibfield{author}{\bibinfo{person}{Zhao Wang}, \bibinfo{person}{Kai Shu}, {and} \bibinfo{person}{Aron Culotta}.} \bibinfo{year}{2021}\natexlab{}.
\newblock \showarticletitle{Enhancing model robustness and fairness with causality: A regularization approach}.
\newblock \bibinfo{journal}{\emph{arXiv preprint arXiv:2110.00911}} (\bibinfo{year}{2021}).
\newblock


\bibitem[Wightman(1998)]%
        {wightman1998lsac}
\bibfield{author}{\bibinfo{person}{Linda~F Wightman}.} \bibinfo{year}{1998}\natexlab{}.
\newblock \bibinfo{booktitle}{\emph{LSAC National Longitudinal Bar Passage Study. LSAC Research Report Series.}}
\newblock \bibinfo{type}{{T}echnical {R}eport}. \bibinfo{institution}{ERIC}.
\newblock


\bibitem[Wright(2006)]%
        {wright2006numerical}
\bibfield{author}{\bibinfo{person}{Stephen~J Wright}.} \bibinfo{year}{2006}\natexlab{}.
\newblock \bibinfo{title}{Numerical optimization}.
\newblock
\newblock


\bibitem[Xie et~al\mbox{.}(2018)]%
        {xie2018differentially}
\bibfield{author}{\bibinfo{person}{Liyang Xie}, \bibinfo{person}{Kaixiang Lin}, \bibinfo{person}{Shu Wang}, \bibinfo{person}{Fei Wang}, {and} \bibinfo{person}{Jiayu Zhou}.} \bibinfo{year}{2018}\natexlab{}.
\newblock \showarticletitle{Differentially private generative adversarial network}.
\newblock \bibinfo{journal}{\emph{arXiv preprint arXiv:1802.06739}} (\bibinfo{year}{2018}).
\newblock


\bibitem[Zafar et~al\mbox{.}(2019)]%
        {zafar2019fairness}
\bibfield{author}{\bibinfo{person}{Muhammad~Bilal Zafar}, \bibinfo{person}{Isabel Valera}, \bibinfo{person}{Manuel Gomez-Rodriguez}, {and} \bibinfo{person}{Krishna~P Gummadi}.} \bibinfo{year}{2019}\natexlab{}.
\newblock \showarticletitle{Fairness constraints: A flexible approach for fair classification}.
\newblock \bibinfo{journal}{\emph{The Journal of Machine Learning Research}} \bibinfo{volume}{20}, \bibinfo{number}{1} (\bibinfo{year}{2019}), \bibinfo{pages}{2737--2778}.
\newblock


\bibitem[Zafar et~al\mbox{.}(2017)]%
        {zafar2017fairness}
\bibfield{author}{\bibinfo{person}{Muhammad~Bilal Zafar}, \bibinfo{person}{Isabel Valera}, \bibinfo{person}{Manuel~Gomez Rogriguez}, {and} \bibinfo{person}{Krishna~P Gummadi}.} \bibinfo{year}{2017}\natexlab{}.
\newblock \showarticletitle{Fairness constraints: Mechanisms for fair classification}. In \bibinfo{booktitle}{\emph{Artificial intelligence and statistics}}. PMLR, \bibinfo{pages}{962--970}.
\newblock


\end{thebibliography}

\end{document}